\documentclass[unnumsec,webpdf,contemporary,large]{oup-authoring-template}%

\usepackage{CJKutf8}
\usepackage{bm}
\usepackage{makecell}
\usepackage{pifont}
\usepackage{color}
\usepackage{xcolor}
\usepackage{graphicx}
\usepackage{url}

\theoremstyle{thmstyleone}%
\theoremstyle{thmstyletwo}%
\theoremstyle{thmstylethree}%

\begin{document}

\journaltitle{Journal Title Here}
\DOI{DOI HERE}
\copyrightyear{2025}
\pubyear{2025}
\access{Advance Access Publication Date: Day Month Year}
\appnotes{Paper}

\firstpage{1}


\title[LabelCoRank: Revolutionizing Long Tail Multi-Label Classification with Co-Occurrence Reranking]{LabelCoRank: Revolutionizing Long Tail Multi-Label Classification with Co-Occurrence Reranking}

\author[1]{Yan Yan}
\author[2]{Junyuan Liu}
\author[3,$\ast$]{Bo-Wen Zhang}

\authormark{Yan yan et al.}

\address[1]{\orgdiv{School of Artificial Intelligence}, \orgname{China University of Mining \& Technology, Beijing}, \orgaddress{\street{Beijing}, \postcode{100083}, \country{China}}}

\address[2]{\orgdiv{School of Artificial Intelligence}, \orgname{China University of Mining \& Technology, Beijing}, \orgaddress{\street{Beijing}, \postcode{100083}, \country{China}}}

\address[3]{ \orgname{Beijing Academy of Artificial Intelligence}, \orgaddress{\street{Beijing}, \postcode{100084}, \country{China}}\ORCID{0000-0002-6384-2104}}


\corresp[$\ast$]{Corresponding author. \href{bwzhang@baai.ac.cn}{bwzhang@baai.ac.cn}}

\received{Date}{0}{Year}
\revised{Date}{0}{Year}
\accepted{Date}{0}{Year}



\abstract{\textbf{Motivation: }Despite recent advancements in semantic representation driven by pre-trained and large-scale language models, addressing long tail challenges in multi-label text classification remains a significant issue. Long tail challenges have persistently posed difficulties in accurately classifying less frequent labels. Current approaches often focus on improving text semantics while neglecting the crucial role of label relationships. \\
\textbf{Results: }This paper introduces LabelCoRank, a novel approach inspired by ranking principles. LabelCoRank leverages label co-occurrence relationships to refine initial label classifications through a dual-stage reranking process. The first stage uses initial classification results to form a preliminary ranking. In the second stage, a label co-occurrence matrix is utilized to rerank the preliminary results, enhancing the accuracy and relevance of the final classifications. By integrating the reranked label representations as additional text features, LabelCoRank effectively mitigates long tail issues in multi-label text classification. Experimental evaluations on popular datasets including MAG-CS, PubMed, and AAPD demonstrate the effectiveness and robustness of LabelCoRank. \\
\textbf{Availability and implementation: }The implementation code is publicly available on  \url{https://github.com/821code/LabelCoRank.}}


\maketitle

\section{Introduction}
\label{Introduction}

Multi-label text classification (MLTC) is a crucial task in natural language processing (NLP) \cite{chen2022survey}. Its goal is to assign appropriate labels to a given text. The exponential growth of textual information on the internet \cite{10.1145/3097983.3098016} has made MLTC increasingly important for extracting valuable insights from vast amounts of text. For instance, in scientific research, the number of published papers is growing exponentially each year. Efficient and cost-effective management of these papers is an urgent problem that needs to be addressed. In the biomedical article retrieval field, MEDLINE, a major component of PubMed maintained by the National Library of Medicine (NLM), uses Medical Subject Headings (MeSH) for indexing. However, MeSH indexing is manually curated, making the process time-consuming and expensive. New articles often take 2 to 3 months to be indexed, with each article costing around \$10 \cite{mao2017mesh}. Essentially, this task can be categorized as multi-label text classification, where each article is assigned several labels. Additionally, MLTC tasks have wide-ranging applications in other fields such as recommendation systems \cite{2018Parabel,JAYALAKSHMI2023103351,Wu_Qiu_Zheng_Zhu_Chen_2024,Wang_Chu_Ouyang_Wang_Hao_Shen_Gu_Xue_Zhang_Cui_Li_Zhou_Li_2024}, sentiment analysis \cite{Jiang2023,GU2023110025}, anomaly detection \cite{Guo_Yang_Liu_Bai_Wang_Li_Zheng_Zhang_Peng_Tian_2024,Chang_Wang_Peng_2024}, rumor detection \cite{Cui_Jia_2024}, and question-answering tasks \cite{Nassiri2023}.

\subsection{Current Situation and Challenges}
However, MLTC faces significant challenges, particularly the long-tail problem. In many datasets, a small number of labels appear frequently (head labels), while the majority of labels are infrequent (tail labels). This long-tail distribution makes it difficult to train effective classifiers, as tail labels lack sufficient training samples. Accurately classifying these less frequent labels remains a significant issue.


Traditional machine learning methods like KNN \cite{ROSEBERRY202110}, SVM \cite{Goudjil2018}, logistic regression \cite{8117734}, and decision trees \cite{9430169} often use feature representations such as Bag-of-Words (BOW) and N-Grams. While these methods have been foundational, they struggle to capture contextual information effectively, leading to limitations in handling complex text data. These traditional methods typically rely on simplistic representations that do not account for the intricate relationships between words and their contexts, resulting in poor performance, especially for rare labels.


Recent advancements in deep learning have led to the development of models based on CNNs and RNNs \cite{chen2015convolutional,10.5555/3060832.3061023,10.1145/3077136.3080834}, which have demonstrated superior feature extraction capabilities. However, these models face challenges in capturing the contextual relationships of words in long texts \cite{bai2018empirical}. For instance, CNNs, while effective in capturing local patterns, struggle with long-range dependencies. RNNs, though better at handling sequential data, often suffer from issues like vanishing gradients, making it hard to learn long-term dependencies effectively. The introduction of attention mechanisms has alleviated some of these issues, with models combining attention mechanisms to improve semantic representations of text and classification performance \cite{yang-etal-2016-hierarchical,NEURIPS2019_9e6a921f,10.1093/bioinformatics/btz142}. Transformer-based models \cite{NIPS2017_3f5ee243} and powerful pre-trained models such as BERT \cite{devlin-etal-2019-bert}, RoBERTa \cite{liu2019roberta}, and others have further enhanced text feature extraction capabilities. These models use self-attention mechanisms to capture relationships between words in a sequence, enabling better handling of long-range dependencies and improving the overall semantic understanding of texts.


Despite these advancements, current models primarily focus on text features and often neglect the relationships among labels and between labels and text. In certain scenarios, relying solely on text features is insufficient for predicting the correct labels, particularly when the text features cannot adequately represent the semantic meaning of the labels. This issue is especially pronounced for tail labels, which lack sufficient training samples and thus are more challenging to associate with text features. For example, when classifying scientific papers based on their abstracts, the short text of the abstracts may not fully convey the semantics of all labels, or the text may be expressed in a more obscure and ambiguous manner. In such cases, it is necessary to utilize the relationships among labels to identify tail labels that are not directly evident from the text semantics. For instance, in medical document classification, the presence of one disease label might imply the likelihood of another co-occurring disease. However, many existing models fail to capture these relationships adequately. Some recent works have tried to incorporate label information into the classification process. For instance, some have used attention mechanisms to explore the relationship between label semantics and text \cite{xiao-etal-2019-label,SUN2024111878}. However, they have not considered the relationships among the labels. Other researchers have employed graph networks to model labels and explore the relationships among labels \cite{ma-etal-2021-label,pal2020multi,Vu2023,li-etal-2022-ligcn,ZENG2024111303}. Yet, in these studies, the expression of the same label information is consistent across different samples, neglecting the fact that label information should exhibit varying importance for different texts.Label relationships are important in multi-label classification tasks because they allow the model to capture co-occurrence patterns and semantic connections between labels, thereby improving prediction accuracy. If label relationships can be further leveraged to dynamically express the important labels for a sample, it would further enhance this property to assist in prediction, particularly for tail labels.

\subsection{Our Contributions}
To address these issues, we propose LabelCoRank, a novel approach inspired by ranking principles. LabelCoRank leverages label co-occurrence relationships to refine initial label classifications through a dual-stage reranking process. The first stage uses initial classification results to form a preliminary ranking. In the second stage, a label co-occurrence matrix is utilized to rerank the preliminary results, enhancing the accuracy and relevance of the final classifications. By integrating the reranked label representations as additional text features, LabelCoRank effectively mitigates long-tail issues in multi-label text classification. The model also incorporates the frequency distribution of labels, allowing it to assign varying importance to labels based on their occurrence in the dataset. Through label embedding and attention mechanisms, LabelCoRank establishes semantic relationships between labels and text features. This dual-stage approach ensures that even infrequent labels, which are typically underrepresented, receive adequate attention during the classification process.

\section{Related Work}

\subsection{Multi-label text classification}

Deep learning has become the mainstream method for addressing the multi-label text classification problem. TextCNN \cite{chen2015convolutional} pioneered the use of convolutional networks to capture local features, integrating these features into text representations through global pooling operations. XML-CNN\cite{10.1145/3077136.3080834} designed a dynamic pooling module to capture richer information from different regions of a document. Compared to convolutional neural networks, recurrent neural networks and attention mechanisms have shown superior performance in capturing contextual associations within texts, making them particularly effective for multi-label text classification tasks.
LSAN\cite{xiao-etal-2019-label} utilizes label semantic information to construct label-specific document representations and employs a self-attention mechanism to measure the contribution of each word to each label. HAN\cite{yang-etal-2016-hierarchical} applies attention mechanisms at both the word and sentence levels, ultimately aggregating these into a text vector to obtain richer text features. AttentionXML\cite{NEURIPS2019_9e6a921f} uses Bi-LSTM(Bidirectional Long Short-Term Memory) to capture long-distance dependencies between words and employs a multi-label attention mechanism to identify the most important parts of the text for each label. Transformer-based models have demonstrated even more powerful feature extraction capabilities, giving rise to a series of widely used pre-trained models such as BERT\cite{devlin-etal-2019-bert}, RoBERTa\cite{liu2019roberta}, and XLNet\cite{NEURIPS2019_dc6a7e65}.Star-Transformer\cite{guo-etal-2019-star} reduces the complexity compared to standard Transformer while retaining the ability to capture local components and long-range dependencies. BERTXML\cite{10.1145/3394486.3403151} introduces the attention mechanism to predict labels based on BERT, achieving good effect improvement. MATCH\cite{10.1145/3442381.3449979}, built on the Transformer architecture, introduces metadata embeddings and uses linear layers to capture their relationship with the text.

However, these methods primarily focus on text feature representation, often neglecting label information. Recently, researchers have begun exploring how to better utilize label information. CorNet\cite{10.1145/3394486.3403151} adds additional linear layer combinations to create the CorNet module, which learns the correlations between labels and can be easily integrated into other models. GUDN\cite{WANG2023161} directly leverages pre-trained models to represent both label semantics and text semantics, using contrastive learning to explore the relationship between labels and text.On the other hand,graph-based methods have emerged as a significant approach to utilizing label information. MAGNET\cite{pal2020multi} uses label embeddings as node features and employs graph attention networks to learn dependencies between labels. LR-GCN\cite{Vu2023} designs a method that utilizes external information from Wikipedia for label embedding, aiming to explore label dependencies and semantics.LDGN\cite{ma-etal-2021-label} combines the statistical label collaboration graph and the dynamic reconstruction graph to form a dual GCN, so as to learn the high-order relationship between labels. LiGCN\cite{li-etal-2022-ligcn} builds a graph for labels and text tokens, and strengthens the edge weight between specific tokens and the corresponding label through the graph convolutional network(GCN). S-GCN\cite{ZENG2024111303} combines text, words, and labels to construct a global graph, considering both semantic associations and global word relationships.By focusing on label information, these methods aim to enhance the multi-label text classification performance, addressing the limitations of traditional text-focused approaches.
\subsection{Label relationship}
\label{sec:format}
In multi-label text classification tasks, the labels of the data set will show different correlations. Different models have different methods of utilizing label relationships. Properly utilizing label relationships is an important research direction. Currently, there are three ways to construct label relationships.

The first direction is to establish label relationships based on the co-occurrence frequency matrix, which is the most intuitive, convenient and commonly used. LACO\cite{zhang-etal-2021-enhancing} constructs two matrices: one is a pairwise label co-occurrence matrix, with 0 and 1 representing infrequent and frequent label co-occurrence respectively, and the other matrix is a conditional label co-occurrence matrix, which indicates whether other labels will co-occur when some labels appear. 
CL-MIL\cite{li-etal-2023-towards-better} will divide labels into three constraint relationships: direct correlation, indirect correlation, and irrelevant. It more fully reflects the relevance strength of the label.

The second direction is to calculate the cosine similarity of label embeddings. When selecting similar labels of a certain label in TailMix\cite{10.1145/3511808.3557632}, the cosine similarity between all one-hop neighbor nodes in the label graph and this node is calculated, and the most similar node is selected.

The third direction is to extract the parent-child relationship between labels. Sometimes,the co-occurrence frequency between labels cannot directly reflect the deep relationship between labels. 
KenMesh\cite{wang-etal-2022-kenmesh} uses a two-layer GCN network to create a graph for the parent label and the child label respectively to reflect the parent-child relationship. However, only considering the parent-child vertical relationship between labels will ignore the horizontal relationship between labels. LGCCN\cite{xu-etal-2021-hierarchical} can capture the vertical dependencies between label levels, model the horizontal correlations, and construct a tree-like label graph without the need for the data set to provide label relationships in advance.

\subsection{Long-tail problem in classification }
In many datasets, labels are not uniformly distributed; rather, a small number of labels (the "head") appear frequently, while the majority (the "tail") occur rarely. This imbalance can lead to poor generalization and biased predictions, particularly for tail labels.Several approaches have been proposed to address the long-tail issue in multi-label classification.

Class rebalancing techniques such as oversampling or undersampling are commonly used to mitigate the effect of label imbalance \cite{4717268} \cite{pmlr-v97-byrd19a} \cite{Zhang2021LearningFS} \cite{ZHAO2024109842}. Square-root sampling \cite{10.1007/978-3-030-01216-8_12}, progressively-balanced sampling \cite{kang2019decoupling}, and adaptive sampling \cite{Wang_2019_ICCV} are also sampling methods that have been proposed in this context. However, these approaches may affect overall representation learning and impact performance on head labels.

Adjusting the loss function is also a common approach to address the long-tail problem. \cite{8953804} rebalances the loss by considering the number of effective samples for each class. \cite{Loss1} proposed a balanced sampling loss function by introducing label co-occurrence probability as a constraint. \cite{huang-etal-2021-balancing} further extended the distribution-balanced Loss into NLP tasks and designed a new loss function called CB-NTR.

Another  direction is the use of transfer learning.HTTN\cite{xiao2021does} transfers meta-knowledge from data-rich head labels to data-scarce tail labels.THETA\cite{10.1145/3643853}, a multi-stage training framework, was proposed for transferring both model-level and feature-level knowledge to improve the performance of tail labels.These methods aim to address the data sparsity issue by enabling the model to learn useful representations even when labeled examples of the tail labels are scarce.

In summary, while a variety of methods have been proposed to handle the long-tail problem in multi-label classification, the challenge remains open. Our work contributes to this area by introducing a re-ranking approach that improves the prediction of tail labels, offering a direction for future research in this domain.

\section{Methodology}
\begin{figure*}[ht]     
	\centering
    \includegraphics[scale=0.17]{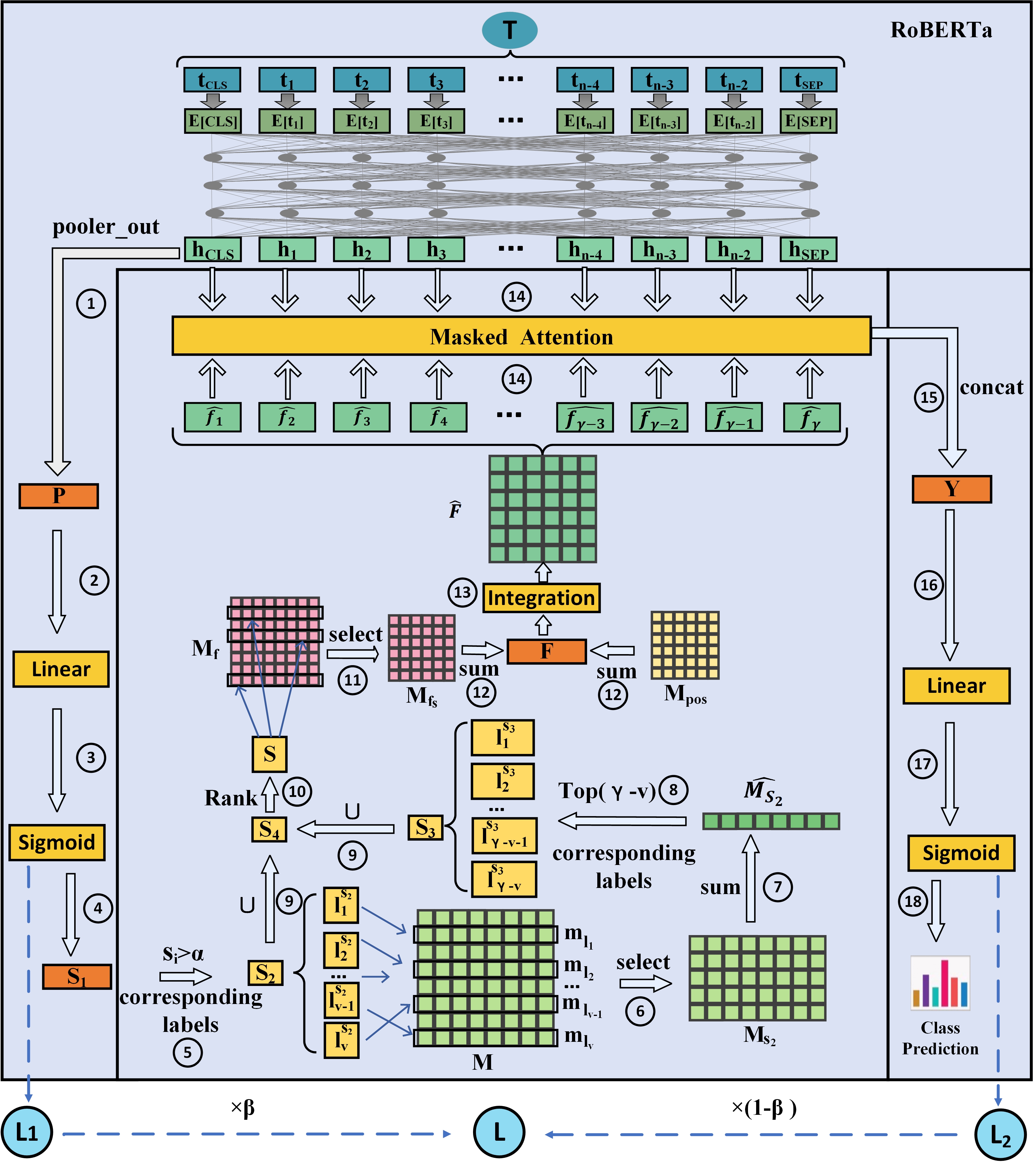} 
	\caption{The architecture of LabelCoRank }   
	\label{FIG:1}
\end{figure*}

In this section, we will first  introduce the overall algorithm flow of LabelCoRank, followed by a detailed explanation of the implementation and motivation of the key component in the model, the Label Reranking with Co-occurrence Relationship method.

\subsection{The whole LabelCoRank pipeline}
\label{sec3.1}
The LabelCoRank model consists of three main modules. The first module uses text features captured by a pre-trained model to obtain an initial classification ranking. In the second module, by leveraging the head labels from the initial classification and combining them with the label frequency co-occurrence matrix and label frequency distribution information from the dataset, Label Reranking is performed. This results in a label feature sequence that incorporates various additional information, which is then used to establish semantic relationships with the text through an attention mechanism. The third module uses these features for the final classification.
Figure \ref{FIG:1} illustrates the overall architecture of the LabelCoRank model. The following three paragraphs will provide a detailed introduction to each of these modules.The numbers from 1 to 18 in the Fig. \ref{FIG:1} indicate the order of the method to help understand the flow of the data.

Initially, the RoBERTa model is employed for feature extraction and initial label prediction based on its text features. The text $T$ is  composed of words $t$, the initial text undergoes tokenization and other preprocessing steps, augmented with special tokens [CLS] and [SEP], resulting in a sequence of $n$ words,$T=\{t_{cls},t_1,t_2,t_3,\dots,t_{n-2},t_{sep}\}$. RoBERTa encodes this text, with the output of its last hidden layer denoted as $H=\{h_{cls},h_1,h_2,h_3,\dots,h_{n-2},h_{sep}\}$.
$h_{cls}$ represents the output of the hidden layer after encoding the [CLS] token and can also be viewed as the representation of the text features after encoding with RoBERTa.The pooled representation of the text is denoted as \(P\), formulated as follows:
\begin{equation}
    H=RoBERTa({W}_{RoBERTa},T)
\end{equation}
\begin{equation}
    P=Tanh(h_{cls}W_1+b_1)
\end{equation}
The full-text features \(P\) encoded by RoBERTa are used for classification, and the sequence of label probabilities is denoted as \( S_1=\{s_1,s_2,\dots s_K\}\), with \(K\) is the number of all labels.
\begin{equation}
    S_1=sigmoid(PW_2+b_2)
\end{equation}
For loss computation, the Binary Cross-Entropy Loss function is used, formulated as follows:
\begin{equation}\label{eq:BCELoss}
    L_{1}=-\frac{1}{K} \sum_{i=1}^{K} \left( \hat{s}_i \log(s_i) + (1 - \hat{s}_i) \log(1 - s_i) \right)
\end{equation}
where \(K\) is the number of all labels, \( \hat{s}_i\) is the true value of the $i$-th label, $s_i$ is the predicted probability value of the $i$-th label.
The prediction loss for this instance is denoted as \(L_1\), which will be combined with the final classification loss \(L_2\) as the ultimate loss L. The motivation for using RoBERTa, a widely-used feature extraction module in NLP that has demonstrated excellent performance, extends to its high accuracy in predicting head labels, although it performs less effectively on tail labels. This discrepancy has inspired us to improve its capability in predicting tail labels while leveraging its strong performance on head labels to use predicted head label information effectively.

Subsequently, the prediction results are combined with the label frequency correlation matrix \( M \) to obtain an extended label sequence, incorporating more relevant labels. This is regarded as the first Label Reranking. The sequence is sorted by frequency to obtain an ordered label sequence \( S \) that includes frequency information. This label sequence is mapped to label features, integrating positional information, and further fitting it through a linear layer. This results in the fused label features \( \widehat{F} = \{ \widehat{f}_1, \widehat{f}_2, \widehat{f}_3, \ldots, \widehat{f}_\gamma \} \), allowing the same label to have different feature representations at different positions in the sequence. This is considered the second Label Reranking.(The specific implementation of \( M \), \( S \) and \(\widehat{F}\) , and a detailed formal explanation of the above content will be provided in the next subsection.) The label feature sequence is then used to perform masked attention learning on the character-level features extracted by RoBERTa, ultimately obtaining features that incorporate expanded label information. The next section will explain how the final label feature sequence containing more relevant information is achieved through Label Reranking, why this process is referred to as Label Reranking, and the motivations behind this approach.

Finally, these features are used for classification. The features that incorporate expanded label information extracted through the multi-head masked attention mechanism, \\ \( \text{MultiHead}(\widehat{f}, H)\), where \( \widehat{f} \in \widehat{F} \), are concatenated to obtain the final feature \( f_{cat} \), as follows:
\begin{equation}
    f_{cat} = Concat(MultiHead(\widehat{f}, H))
\end{equation}
The features are mapped to the corresponding probabilities of \( K \) labels through a linear transformation and a sigmoid function.
\begin{equation}
    Y = sigmoid(f_{cat} W_5 + b_5)
\end{equation}
The binary cross-entropy loss function for \( Y =\{y_1,y_2,\dots y_K\} \) is used as the loss function for the second prediction, resulting in the loss \( L_2 \). The final loss is defined as:
\begin{equation}\label{eq:BCELoss2}
    L_{2}=-\frac{1}{K} \sum_{i=1}^{K} \left( \hat{y}_i \log(y_i) + (1 - \hat{y}_i) \log(1 - y_i) \right)
\end{equation}
\begin{equation}
    L = \beta \times L_{1} + (1 - \beta) \times L_{2}
\end{equation}
where  \(K\) is the number of all labels, \( \hat{y}_i\) is the true value of the $i$-th label, $y_i$ is the predicted probability value of the $i$-th label and \( \beta \) is a hyperparameter used to balance the influence of the two losses on the final result.

\subsection{Label Reranking with Co-occurance Relationship}
\label{sec3.2}
In the first stage, an initial ranking is formed using the preliminary classification results. In the second stage, we re-rank these results to integrate various additional information beyond text features, helping to address the long-tail problem in classification. This section will specifically introduce the method of Label Reranking with Co-occurrence Relationship.

After the initial prediction, the first predicted label probability sequence \( S_1 \) is obtained. A threshold hyperparameter \( \alpha \) is used to select the parts of \( S_1 \) where the label prediction probabilities are greater than \( \alpha \), and the corresponding labels are taken as the label sequence \( S_2 \).  This aims to obtain labels more relevant to the text as predicted by RoBERTa, thereby reducing the introduction of irrelevant noise.

For the co-occurrence frequency matrix \( M \), its contents are defined as follows: for a dataset with \( K \) labels, \( m_{i,j} \) represents the frequency with which the \( j \) label appears when the \( i \) label is present. \( m_i \) is the co-occurrence frequency sequence of all labels corresponding to the \( i \) label, that is, \( m_i = \{m_{i,1}, m_{i,2}, ..., m_{i,k}\} \). Therefore, for the label sequence \( S_2 \) with \( V \) labels, \( S_2 = \{l_{1}, l_{2}, \ldots, l_{v}\} \), the label frequency sequence set is \( M_{S_2} = \{m_{l_1}, m_{l_2}, \ldots, m_{l_v}\} \). Summing the label frequency sequences yields \( \widehat{M_{s_2}} \), defined as follows:
\begin{equation}
\widehat{M_{s_2}} = \sum_{v=1}^{V} m_{l_v} = \left \{ \sum_{v=1}^{V} m_{l_v,1}, \sum_{v=1}^{V} m_{l_v,2}, \ldots, \sum_{v=1}^{V} m_{l_v,k} \right \}
\end{equation}
A hyperparameter \( \gamma \) is used to control the desired length of the label sequence. The corresponding labels from \( \widehat{M_{s_2}} \) are obtained, and the top \( \gamma - V \) labels that are not part of \( S_2 \) are selected based on frequency in descending order to form the label sequence \( S_3 \). \( S_2 \cup S_3 \) is used as the label sequence \( S_4 \), which is more relevant to each sample:
\begin{equation}
S_4 = S_2 \cup S_3
\end{equation}

This is regarded as the first Label Reranking. In terms of effectiveness, using the co-occurrence frequency matrix allows samples to expand to include more relevant labels, thereby obtaining more related information. However, from another perspective, through the label co-occurrence matrix, important information in the initial label probability sequence that might have been overlooked is reordered and reprioritized based on co-occurrence information.

\( S_4 \) is then sorted based on the label frequency distribution across the entire dataset, placing high-frequency labels at the beginning of the sequence and low-frequency labels at the end. This results in a frequency-integrated label sequence \( S \):
\begin{equation}
S = Rank(S_4)
\end{equation}

A trainable label position feature matrix \( M_{\text{pos}} \in \mathbb{R}^{\gamma \times \delta} \) and a label feature matrix \( M_{\text{f}} \in \mathbb{R}^{K \times \delta} \) are designed, where \( \delta \) is the size of the RoBERTa hidden layer. Using \( S \), the corresponding label features from \( M_{\text{f}} \) are selected and added to \( M_{\text{pos}} \) to obtain the feature representation \( F \) with integrated label position information. A feedforward neural network module is then used to further integrate the position information and label information in \( F \), represented as follows:
\begin{equation}\label{F}
\hat{F} = ReLU(Drop(F W_3 + b_3)) W_4 + b_4
\end{equation}

This is considered the second Label Reranking. By sorting labels according to their distribution in the dataset, label frequency distribution information is incorporated, and all relevant labels are re-ranked. This effectively organizes the highly semantically relevant head labels initially predicted by RoBERTa and the highly correlated labels obtained through co-occurrence relationships.

The design has two main motivations. First, by reranking the labels based on the frequency distribution in the dataset, high-frequency labels are placed at the beginning of the sequence. This way, the selected label feature information sequence also contains the importance information of frequency. During subsequent network training using label feature information, the neural network gives different levels of attention to the information at the beginning and end of the sequence, thus integrating this information into the model. Second, the position feature matrix is designed because different samples have different label sequences \( S \), and the order of \( S \) contains label frequency information. For the same label, its meaning should be different depending on its position in different label sequences. After sorting and filtering, the labels that appear earlier in the sequence are more important, whereas the labels that appear later are less important. Therefore, the same label in different positions within the sequence supplements the corresponding label with feature information specific to its position.

In the end, the label feature sequence \( \widehat{F} = \{ \widehat{f}_1, \widehat{f}_2, \ldots, \widehat{f}_\gamma \} \), combined with a multi-head masked attention mechanism, is used to extract information from the RoBERTa-encoded text that is more relevant to the label information.
\begin{equation}
    MultiHead(\widehat{f},H)=Concat(head_1,head_2,head_3...head_\eta)
\end{equation}
\begin{equation}
head_i=Attention(Q_i, K_i, V_i) = softmax(\frac{Q_i K_i^T}{\sqrt{d_k}} + M) V_i
\end{equation}

where \(Q_i = \widehat{f}W_i^Q,K_i = H W_i^K,V_i = H W_i^V  \quad for\; i \in[1, \eta], \widehat{f} \in \widehat{F} \) .
\(\eta\) is the number of heads . \( W_i^Q \), \( W_i^K \), and \( W_i^V \) are the  matrices parameters to be learned for the \( i \)-th head.  \( d_k \) is the dimension of the keys,  and \( M \) is the attention mask that is added to the scores.

The above outlines the entire process of Label Reranking with Co-occurrence Relationship. Its purpose is to integrate various additional label information through Label Reranking to address the long-tail problem in multi-label text classification. In the following comparative and ablation experiments, we will explore the effectiveness of this method and verify the design of LabelCoRank.

\section{Experiments}
\subsection{Datasets}
The proposed model was evaluated on three publicly available datasets. Table \ref{tab:Dataset statistics} presents the statistics of these three datasets. 

     \textbf{MAG-CS}\cite{10.1162/qss_a_00021}: The dataset consists of 705,407 papers from the Microsoft Academic Graph (MAG), selected from 105 prominent CS conferences held between 1990 and 2020. It comprises 15,808 unique labels, providing a comprehensive collection of scientific literature in the field.
     
     \textbf{PubMed}\cite{Lu2011-gz}: The dataset comprises 898,546 papers sourced from PubMed, representing 150 leading medical journals from 2010 to 2020. It includes 17,963 labels corresponding to MeSH terms, offering valuable insights into biomedical research.
     
     \textbf{AAPD}\cite{yang-etal-2018-sgm}: The dataset comprises English abstracts of computer science papers sourced from arxiv.org, each paired with relevant topics. In total, it includes 55,840 abstracts spanning various related disciplines.
     
\begin{table}[htbp]
\caption{Dataset statistics.$N_{\textit{trn}}$ and $N_{\textit{tst}}$ represent the number of documents in the training set and test set, respectively. $D$ represents the vocabulary size of all documents. $L$ represents the number of labels. $L_{\textit{avg}}$ represents the average number of labels per document, and $W_{\textit{avg}}$ represents the average number of words per document.}
\renewcommand{\arraystretch}{1.07}
\centering

\begin{tabular}{lcccccc}
\hline
Datasets  & $N_{\textit{trn}}$ & $N_{\textit{tst}}$ & $D$ & $L_n$ & $L_{\textit{avg}}$ & $W_{\textit{avg}}$\\ 
\hline
 MAG-CS & 564340 & 70534 & 425345 & 15809 & 5.60 & 126.33\\
 PubMed & 718837 & 89855 & 776975 & 17963 & 7.78 & 198.97\\
 AAPD & 54840 & 1000 & 69399 & 54 & 2.41 & 163.43\\
\hline
\end{tabular}

\label{tab:Dataset statistics}
\end{table}

\subsection{Baselines and Evaluation metrics}
\subsubsection{Baselines}
The following comparison methods were selected. They are categorized into three types: methods based on traditional neural networks, methods based on Transformers, and methods similar to ours that also utilize pre-trained models and incorporate label information.
\begin{itemize}
    \item \textbf{XML-CNN}\cite{10.1145/3077136.3080834} utilizes a dynamic max pooling scheme to capture richer information from different regions of the document, adopts a binary cross-entropy loss function to handle multi-label problems, and introduces hidden bottleneck layers to obtain better document representation and reduce model size.
    \item \textbf{MeSHProbeNet}\cite{10.1093/bioinformatics/btz142} is an end-to-end deep learning model designed for MeSH indexing, which assigns MeSH terms to MEDLINE citations. It won the first place in the latest batch of Task A in the 2018 BioASQ challenge.
    \item \textbf{AttentionXML}\cite{NEURIPS2019_9e6a921f} is a label tree-based deep learning model designed for Extreme Multi-label Text Classification (XMTC). It introduces two key features: a multi-label attention mechanism that captures the relevant text parts of each label, and a shallow and wide probabilistic label tree that can efficiently handle millions of labels.
    \item \textbf{Transformer}\cite{NIPS2017_3f5ee243} is a network architecture, the first sequence transformation model based entirely on the attention mechanism. It replaces the commonly used cyclic layer with a multi-head self-attention mechanism and abandons the common cyclic and CNN structures.
    \item \textbf{Star-Transformer}\cite{guo-etal-2019-star} is a lightweight alternative to Transformer for NLP tasks. It uses a star topology to reduce the complexity from quadratic to linear and solve the problem of heavy computing requirements.
    \item \textbf{RoBERTa}\cite{yinhan2019roberta} is an advanced variant of BERT, designed to improve the pretraining process by optimizing key aspects such as training data size, batch size, and training time. 
    \item \textbf{BertXML}\cite{10.1145/3394486.3403151} is a customized version of BERT designed specifically for XMTC. It overcomes the limitation of BERT's single [CLS] token by incorporating multiple [CLS] tokens at the start of each input sequence.
    \item \textbf{MATCH}\cite{10.1145/3442381.3449979} learns improved text and metadata representations by jointly embedding them in the same space, enabling higher-order interactions between words and metadata.
\item \textbf{LiGCN}\cite{li-etal-2022-ligcn} introduced an interpretable graph convolutional network model that models tokens and labels as nodes in a heterogeneous graph. It calculates the cosine similarity between label embeddings to capture the relationships among labels.
    \item \textbf{GUDN}\cite{WANG2023161} leverages label semantics and deep pre-trained models, incorporating a label reinforcement strategy for fine-tuning to enhance classification performance. The model exhibits sensitivity to label semantics and shows remarkable efficacy on datasets with semantically rich labels.
\end{itemize}
\subsubsection{Evaluation metrics}
In datasets for multi-label text classification with extensive label sets, each text typically associates with only a few labels (see Table \ref{tab:Dataset statistics}). The distribution of labels is uneven and sparse. Hence, it's customary to assess classification quality by generating a concise ranked list of potentially relevant labels for each test document. We follow the research by Zhang et al.\cite{10.1145/3442381.3449979},  adhering to two rank-based evaluation metrics: precision at top $k$ (P@$k$) and normalized Discounted Cumulative Gain at top $k$ (NDCG@$k$), where $k$ = 1, 3, 5. For a document $d$, let $\bm{y}_{\textit{d}} \in \{0, 1\} ^ {\left| \mathcal{L} \right|}$ be its ground truth label vector and rank($i$) be the index of the $i$-th highest predicted label according to the re-ranker.
\begin{equation}
P@k=\frac{1}{k}\sum_{i=1}^{k}\bm{y}_{d,rank(i)}
\end{equation}

\begin{equation}
DCG@k=\sum_{i=1}^{k}\frac{\bm{y}_{d,rank(i)}}{\log_{2}{(i+1)}}
\end{equation}

\begin{equation}
NDCG@k = \frac{DCG@k}{\sum_{i=1}^{\min(k,\left||\bm{y}_{d}\right||_{0})}\frac{1}{\log_{2}{(i+1)}}}
\end{equation}

\subsection{Experimental settings}
All experimental environments were consistent. Experiments were conducted on a computer equipped with an Nvidia RTX 4090 GPU and 128 GB of RAM. All comparative methods using pre-trained models, as well as our approach, utilized the RoBERTa pre-trained model as a feature extractor. The hyperparameters for all comparative methods were set according to their original papers or code. The AdamW optimizer was employed with a learning rate of 1e-5, sentence truncation set to 512, and batch size of 16. The threshold hyperparameter, $\alpha$, was set to 0.3, and the hyperparameters for the loss function weight, $\beta$, was set to 0.3, 0.3, and 0.25 for the MAG-CS, PubMed, and AAPD datasets, respectively. The number of selected labels for these datasets \(\gamma\) was 30, 35, and 20, respectively.
\subsection{Experimental results}
\subsubsection{Performance on the complete datasets}
Tables \ref{tab:Resutl on MAG}, \ref{tab:Resutl on PubMed}, and \ref{tab:Resutl on AAPD} summarize the results of different models on the MAG-CS, PubMed, and AAPD datasets, respectively.

In the MAG-CS dataset (Table \ref{tab:Resutl on MAG}), the model outperforms all baselines on almost all metrics, except for P@1, where it lags behind MATCH by 0.0072. However, for P@3, P@5, NDCG@3 and NDCG@5, the model improves over MATCH by 0.0047, 0.0131, 0.0007 and 0.0079, respectively. This indicates an enhanced prediction performance for tail labels. This improvement is attributed to the introduction of a significant amount of relevant label information, which boosts the prediction of tail labels but slightly dilutes the focus on head labels. Additionally, MATCH utilizes extra meta-information that LabelCoRank does not, which could explain the differences for P@1.

\begin{table}[htbp]
\caption{Results on MAG. Following the experiments from the MATCH\cite{10.1145/3442381.3449979}, the average of three trials was taken for comparison. The best experimental result on the MAG dataset is indicated in bold. }
\renewcommand{\arraystretch}{1.07}
\centering
\resizebox{\linewidth}{!}{
    \begin{tabular}{lccccc} \Xhline{1.1pt}
         Model&  P@1&  P@3&  P@5&  NDCG@3& NDCG@5\\ \hline 
         XML-CNN\cite{10.1145/3077136.3080834}&  0.8656&  0.7028&  0.5756&  0.7842& 0.7407\\
         MeSHProbeNet\cite{10.1093/bioinformatics/btz142}&  0.8738&  0.7219&  0.5927&  0.8020& 0.7588\\
         AttentionXML\cite{NEURIPS2019_9e6a921f}&  0.9035&  0.7682&  0.6441&  0.8489& 0.8145\\  \hline
         Transformer\cite{NIPS2017_3f5ee243}&  0.8805&  0.7327&  0.6024&  0.8129& 0.7703\\ 
         Star-Transformer\cite{guo-etal-2019-star}&  0.8569&  0.7089&  0.5853&  0.7876& 0.7486\\
         RoBERT\cite{yinhan2019roberta}a&  0.8862&  0.7475&  0.6258&  0.8277& 0.7932\\ 
         BERTXML\cite{10.1145/3394486.3403151}&  0.9011&  0.7532&  0.6238&  0.8355& 0.7954\\ 
         MATCH\cite{10.1145/3442381.3449979}&  \textbf{0.9190}&  0.7763&  0.6457&  0.8610& 0.8223\\ \hline 
         RoBERTa-LiGCN\cite{li-etal-2022-ligcn}&  -&  -&  -&  -& -\\ 
         GUDN\cite{WANG2023161}&  0.8757&  0.7354&  0.6173&  0.8137& 0.7805\\ 
 LabelCoRank& 0.9118& \textbf{0.7810}&\textbf{ 0.6588}& \textbf{0.8617}&\textbf{0.8302}\\ \Xhline{1.1pt}
\end{tabular}}

\label{tab:Resutl on MAG}
\end{table}

In the PubMed dataset (Table \ref{tab:Resutl on PubMed}), LabelCoRank achieves the best performance across all metrics. It is worth noting that LabelCoRank demonstrates a significantly better performance than MATCH on the PubMed dataset compared to the MAG dataset. The improvements in P@3, P@5, NDCG@3, and NDCG@5 all exceed 2.8 percentage points, suggesting a substantial improvement in predicting difficult-to-predict tail labels. The PubMed dataset has the highest average number of labels per instance, which likely enhances the relevance of supplementary label information, leading to superior performance.
\begin{table}
\caption{Results on PubMed. Following the experiments from the MATCH\cite{10.1145/3442381.3449979}, the average of three trials was taken for comparison. The best experimental result on the PubMed dataset is indicated in bold.  }
\renewcommand{\arraystretch}{1.07}
    \centering
    \resizebox{\linewidth}{!}{
    \begin{tabular}{lccccc} \Xhline{1.1pt}
         Model&  P@1&  P@3&  P@5&  NDCG@3& NDCG@5\\ \hline 
         XML-CNN\cite{10.1145/3077136.3080834}&  0.9084&  0.7182&  0.5857&  0.7790& 0.7075\\ 
         MeSHProbeNet\cite{10.1093/bioinformatics/btz142}&  0.9135&  0.7224&  0.5878&  0.7836& 0.7109\\ 
         AttentionXML\cite{NEURIPS2019_9e6a921f}&  0.9125&  0.7414&  0.6169&  0.7979& 0.7341\\  \hline
         Transformer\cite{NIPS2017_3f5ee243}&  0.8971&  0.7299&  0.6003&  0.7867& 0.7178\\ 
         Star-Transformer\cite{guo-etal-2019-star}&  0.8962&  0.6990&  0.5641&  0.7612& 0.6869\\
         RoBERTa\cite{yinhan2019roberta}&  0.9178&  0.7343&  0.6037&  0.7941& 0.7254\\ 
         BERTXML\cite{10.1145/3394486.3403151}&  0.9144&  0.7362&  0.6032&  0.7949& 0.7247\\ 
         MATCH\cite{10.1145/3442381.3449979}&  0.9168&  0.7511&  0.6199&  0.8072& 0.7395\\ \hline 
         RoBERTa-LiGCN\cite{li-etal-2022-ligcn}&  -&  -&  -&  -& -\\ 
         GUDN\cite{WANG2023161}&  0.9241&  0.7760&  0.6524&  0.8285& 0.7683\\ 
 LabelCoRank& \textbf{0.9243}& \textbf{0.7794}& \textbf{0.6571}& \textbf{0.8315}&\textbf{0.7729}\\ \Xhline{1.1pt}
    \end{tabular}} 

\label{tab:Resutl on PubMed}
\end{table}

In the AAPD dataset (Table \ref{tab:Resutl on AAPD}), LabelCoRank also achieves the best performance across all metrics. Despite the average number of labels per document being only 2.41, the total number of labels is 54, allowing most label features to participate in the model's computations. LabelCoRank effectively learns and distinguishes relevant from irrelevant label information without introducing excessive noise from additional labels. This demonstrates the model's robustness even in tasks with fewer labels.

\begin{table}[htb]
\caption{Results on AAPD. Following the experiments from the LiGCN\cite{li-etal-2022-ligcn}, the best experimental results were taken for comparison. The best experimental result on the AAPD dataset is indicated in bold.  }
\renewcommand{\arraystretch}{1.07}
    \centering
    \resizebox{\linewidth}{!}{
    \begin{tabular}{lccccc} \Xhline{1.1pt}
         Model&  P@1&  P@3&  P@5&  NDCG@3& NDCG@5\\ \hline 
         XML-CNN\cite{10.1145/3077136.3080834}&  0.8140&  0.5903&  0.4064&  0.7771& 0.8187\\ 
         MeSHProbeNet\cite{10.1093/bioinformatics/btz142}&  0.8460&  0.6020&  0.4130&  0.7969& 0.8384\\  
         AttentionXML\cite{NEURIPS2019_9e6a921f}&  0.8500&  0.6110&  0.4180&  0.8081& 0.8483\\ \hline 
         Transformer\cite{NIPS2017_3f5ee243}&  0.7820&  0.5687&  0.3930&  0.7456& 0.7853\\  
         Star-Transformer\cite{guo-etal-2019-star} &  0.7790&  0.5697&  0.3976&  0.7471& 0.7917\\
         RoBERTa\cite{yinhan2019roberta}&  0.8190&  0.6080&  0.4096&  0.7806& 0.8104\\ 
         BERTXML\cite{10.1145/3394486.3403151}&  0.8140&  0.5847&  0.3972&  0.7737& 0.8093\\  
         MATCH\cite{10.1145/3442381.3449979}&  -&  -&  -&  -& -\\ \hline 
         RoBERTa-LiGCN\cite{li-etal-2022-ligcn}&  0.8250&  0.6126&  0.4138&  0.8039& 0.8383\\  
         GUDN\cite{WANG2023161}&  0.8260&  0.6023&  0.4136&  0.7911& 0.8335\\ 
 LabelCoRank& \textbf{0.8540}& \textbf{0.6227}& \textbf{0.4188}& \textbf{0.8188}&\textbf{0.8526}\\ \Xhline{1.1pt}
    \end{tabular}}

\label{tab:Resutl on AAPD}
\end{table}

From these experiments, it is evident that models based on the Transformer architecture outperform traditional CNN and RNN-based models. When comparing models that also use pre-trained models and incorporate label information, namely RoBERTa-LiGCN and GUDN, LabelCoRank consistently outperforms them. GUDN's performance on the MAG dataset is weaker than on PubMed, likely because it relies directly on pre-trained models to extract label features, which are presented as IDs in MAG, making it difficult to capture label text features effectively. However, LabelCoRank allows the model to learn the feature representation of each label independently, without relying on label text features.
\subsubsection{ Performance Comparison of Head and Tail Labels}

\begin{figure}[ht]     
	\centering
    \includegraphics[scale=0.2]{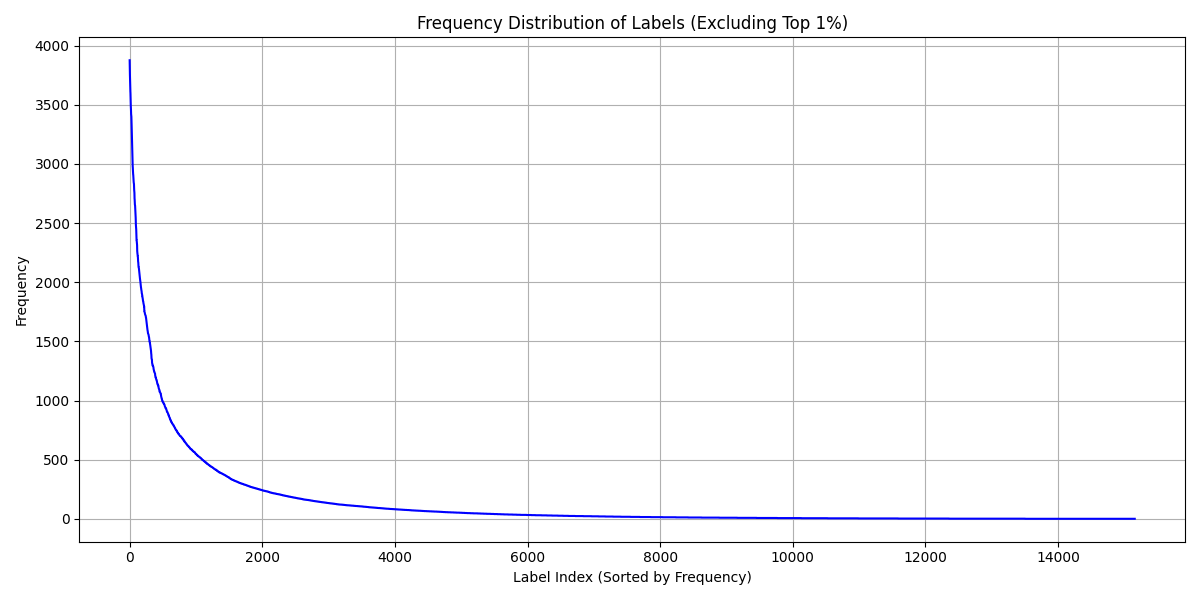} 
	\caption{Frequency Distribution of Labels on PubMed}   
	\label{FIG:2}
\end{figure}

To verify the improvement of LabelCoRank in tail label prediction, we selected two subsets from the PubMed dataset for the experiment. The first subset is the head label dataset, which consists of samples where the real labels are entirely formed by the top 10\% most frequent head labels, totaling 226,629 samples. The second subset is the tail label dataset, which consists of samples where the top labels appear no more than twice in the real labels, totaling 93,364 samples. Figure \ref{FIG:2} shows the label distribution of the PubMed dataset after excluding the top 1\% frequent labels (the distribution of the top 1\% labels is not displayed in the figure to prevent the graph from becoming too "L"-shaped). The top 10\% most frequent labels are considered head labels, while the remaining 90\% are considered tail labels. Since the dataset consisting entirely of tail label samples only contains 3,803 samples, we selected a subset consisting of samples where the top labels appear no more than twice in the real labels.

\begin{table}[htb]
\caption{Results on PubMed head label dataset}
\renewcommand{\arraystretch}{1.07}
    \centering
    \resizebox{\linewidth}{!}{
     \begin{tabular}{lccccc} \Xhline{1.1pt}
         Model&  P@1&  P@3&  P@5&  NDCG@3& NDCG@5\\ \hline 
         XML-CNN\cite{10.1145/3077136.3080834}&  0.8692&  0.6318&  0.4914&  0.7409& 0.6969\\
         MeSHProbeNet\cite{10.1093/bioinformatics/btz142}&  0.9168&  0.7183&  0.5665&  0.8233& 0.7806\\
         AttentionXML\cite{NEURIPS2019_9e6a921f}&  0.9122&  0.7111&  0.5680&  0.8155& 0.7784\\  \hline
          Transformer\cite{NIPS2017_3f5ee243}&  0.8686&  0.6371&  0.4914&  0.7426& 0.6872\\ 
         Star-Transformer\cite{guo-etal-2019-star}&  0.8709&  0.6341&  0.4915&  0.7404& 0.6871\\
         RoBERTa\cite{yinhan2019roberta}&  0.8960&  0.6633&  0.5162&  0.7738& 0.7284\\ 
         BERTXML\cite{10.1145/3394486.3403151}&  0.8854&  0.6606&  0.5116&  0.7675& 0.7199\\ 
         MATCH\cite{10.1145/3442381.3449979}&   -&  -&  -&  -& -\\ \hline 
         RoBERTa-LiGCN\cite{li-etal-2022-ligcn}&  -&  -&  -&  -& -\\ 
         GUDN\cite{WANG2023161}&  0.9072&  0.7040&  0.5558&  0.8089& 0.7671\\ 
 LabelCoRank& \textbf{0.9178}& \textbf{0.7240}&\textbf{ 0.5752}& \textbf{0.8284}&\textbf{0.7887}\\ \Xhline{1.1pt}
\end{tabular}}
    \label{tab:PubMed head label}
\end{table}

\begin{table}[htb]
\caption{Results on PubMed tail label dataset}
\renewcommand{\arraystretch}{1.07}
    \centering
    \resizebox{\linewidth}{!}{
     \begin{tabular}{lccccc} \Xhline{1.1pt}
         Model&  P@1&  P@3&  P@5&  NDCG@3& NDCG@5\\ \hline 
         XML-CNN\cite{10.1145/3077136.3080834}&  0.7509&  0.4319&  0.3017&  0.5287& 0.5142\\
         MeSHProbeNet\cite{10.1093/bioinformatics/btz142}&  0.7925&  0.4927&  0.3450&  0.5910& 0.5753\\
         AttentionXML\cite{NEURIPS2019_9e6a921f}&  0.7997&  0.4878&  0.3458&  0.5856& 0.5727\\  \hline
         Transformer\cite{NIPS2017_3f5ee243}&  0.7784&  0.4640&  0.3188&  0.5638& 0.5347\\ 
         Star-Transformer\cite{guo-etal-2019-star}&  0.7625&  0.4511&  0.3122&  0.5489& 0.5227\\
         RoBERTa\cite{yinhan2019roberta}&  0.7739&  0.4586&  0.3141&  0.5516& 0.5278\\ 
         BERTXML\cite{10.1145/3394486.3403151}&  0.8014&  0.5213&  0.3629&  0.6170& 0.5974\\ 
         MATCH\cite{10.1145/3442381.3449979}&   -&  -&  -&  -& -\\ \hline 
         RoBERTa-LiGCN\cite{li-etal-2022-ligcn}&  -&  -&  -&  -& -\\ 
         GUDN\cite{WANG2023161}&  0.8192&  0.5306&  0.3744&  0.6279& 0.6136\\ 
 LabelCoRank& \textbf{0.8382}& \textbf{0.5593}&\textbf{ 0.3940}& \textbf{0.6584}&\textbf{0.6431}\\  \Xhline{1.1pt}

\end{tabular}}
    \label{tab:PubMed Tail label}
\end{table}

On the head label dataset (Table \ref{tab:PubMed head label}), LabelCoRank outperforms the current state-of-the-art models across all evaluation metrics.LabelCoRank shows an improvement of nearly 0.6\% and 0.9\% over the second-best method in P@3 and P@5, respectively, and also achieves gains in the NDCG metrics.On the tail label dataset (Table \ref{tab:PubMed Tail label}), the advantage of LabelCoRank is more pronounced. Compared to other models, LabelCoRank achieves the best results in P@1 (0.8382), P@3 (0.5593), P@5 (0.3940), NDCG@3 (0.6584), and NDCG@5 (0.6431). For instance, P@3 and P@5 are improved by approximately 2.9\% and 2.0\% compared to GUDN, while NDCG@3 and NDCG@5 are improved by around 3.0\%. These results clearly demonstrate that LabelCoRank significantly improves the prediction accuracy  for tail labels.

\subsubsection{Comparison with GPT-4o}

\begin{table}[htb]
\caption{Results on PubMed. RoBERTa* denotes the result where all correct results are selected from the top 10 predictions of RoBERTa.}
\renewcommand{\arraystretch}{1.07}
    \centering
    \resizebox{\linewidth}{!}{
     \begin{tabular}{lccccc} \Xhline{1.1pt}
         Model&  P@1&  P@3&  P@5&  NDCG@3& NDCG@5\\\hline
         RoBERTa*&  0.9935&  0.9046&  0.7484&  0.9474& 0.8719\\ \hline
          RoBERTa&  0.9178&  0.7343&  0.6037&  0.7941& 0.7254\\ 
          RoBERTa+GPT-4o&  0.7990&  0.6343&  0.5304&  0.6828& 0.6271\\ 
         LabelCoRank& \textbf{0.9243}& \textbf{0.7794}& \textbf{0.6571}& \textbf{0.8315}&\textbf{0.7729}\\\Xhline{1.1pt} 
\end{tabular}}
    \label{tab:PubMed with GPT}
\end{table}

In this section, we compare LabelCoRank with GPT-4o, a renowned advanced large language model, on the PubMed dataset. Due to the prompt length limitation, we are unable to provide the GPT-4o with all the labels, thus preventing us from fully leveraging the capabilities of the GPT-4o. Instead, we use the GPT-4o to rerank the top ten predictions from RoBERTa and select the five most relevant labels. Viewed from another angle, this approach is a secondary enhancement of RoBERTa’s predictions by GPT-4o, which is conceptually similar to what LabelCoRank does. The following prompt was used in the experiment: ``Given the text '\{data\}', and the following labels: \{labels\}, with corresponding indices \{preList\}, please select the five most relevant labels for the text and return the corresponding labels and indices in Python list format. For example: ['A', 'B', 'C', 'D', 'E'] and [1, 6, 3, 8, 4]. If there are fewer than 5 results, add the ones you find most relevant to make up the difference. No explanation is needed, just the relevant label list and its index list.$\backslash$n". Additionally, we calculated the metric that indicates while all the correct labels be selected from RoBERTa’s top ten predictions, which helps establish an upper bound to avoid the result being impacted by the lack of correct labels within RoBERTa's top 10 predictions. The results after GPT-4o reranking were parsed and calculated, and the findings are in table \ref{tab:PubMed with GPT}.

The results show that our method outperforms GPT's selection based on RoBERTa’s top 10 predictions. Moreover, we found that GPT’s performance actually deteriorated after reranking, which may suggest that large language model struggles to adapt well to this task.

\subsection{Ablation study}
From the analysis of the above experiments, it is evident that LabelCoRank effectively improves label prediction performance by incorporating label information, particularly enhancing the accuracy of predicting tail labels in datasets with a large number of labels. We conducted ablation experiments to verify the effectiveness of the modules used in LabelCoRank for integrating label information. Four design elements in LabelCoRank need to be validated for their effectiveness: label selection, label sequence ranking, incorporation of positional information, and selection of the number of labels.
\subsubsection{Effects of label selection,label sequence ranking position information and all all reranking parts}
\begin{table}[htb]
\caption{Results of ablation experiments on MAG-CS}
\renewcommand{\arraystretch}{1.07}
    \centering
    \resizebox{\linewidth}{!}{
    \begin{tabular}{lccccc} \Xhline{1.1pt} 
         Model&  P@1&  P@3&  P@5&  NDCG@3& NDCG@5\\ \hline
         w/o correlation matrix&  0.9031&  0.7680&  0.6442&  0.8483& 0.8139\\
         w/o label ranking&  0.9056&  0.7725&  0.6497&  0.8527& 0.8197\\ 
         w/o  position information&  0.9039&  0.7692&  0.6452&  0.8498& 0.8156\\ 
         w/o  all reranking parts &  0.8997&  0.7578&  0.6342&  0.8402& 0.8062\\ 
         LabelCoRank&  \textbf{0.9118}&  \textbf{0.7811}&  \textbf{0.6591}&  \textbf{0.8620}& \textbf{0.8305}\\ \Xhline{1.1pt}
    \end{tabular}}
    \label{tab:ablationMAG-CS}
\end{table}

\begin{table}
    \caption{Results of ablation experiments on PubMed }
\renewcommand{\arraystretch}{1.07}
    \centering
    \resizebox{\linewidth}{!}{
    \begin{tabular}{lccccc} \Xhline{1.1pt} 
         Model&  P@1&  P@3&  P@5&  NDCG@3& NDCG@5
\\ \hline
w/o correlation matrix&  0.9251&  0.7700&  0.6442&  0.8243& 0.7623\\
         w/o label ranking&  \textbf{0.9287}&  0.7761&  0.6519&  0.8298& 0.7694\\ 
         w/o position information&  0.9259&  0.7699&  0.6441&  0.8242& 0.7621\\  
         w/o  all reranking parts&  0.9236&  0.7608&  0.6337&  0.8165& 0.7527\\
         LabelCoRank&  0.9241&  \textbf{0.7813}&  \textbf{0.6593}&  \textbf{0.8330}& \textbf{0.7749}\\ \Xhline{1.1pt}
    \end{tabular}}

        \label{tab:ablationPubMed}
\end{table}

\begin{table}[htb]
\caption{Results of ablation experiments on AAPD }
\renewcommand{\arraystretch}{1.07}

    \centering
    \resizebox{\linewidth}{!}{
    \begin{tabular}{lccccc} \Xhline{1.1pt} 
         Model&  P@1&  P@3&  P@5&  NDCG@3& NDCG@5
\\ \hline
         w/o correlation matrix&  0.8470&  0.6193&  \textbf{0.4204}&  0.8129& 0.8509\\
        w/o label ranking&  0.8490&  0.6193&  0.4192&  0.8145& 0.8515\\  
         w/o position information&  0.8480&  0.6160&  0.4170&  0.8138& 0.8504\\ 
        w/o  all reranking parts&  0.8340&  0.6132&  0.4156&  0.8077& 0.8423\\
         LabelCoRank&  \textbf{0.8540}&  \textbf{0.6227}&  0.4188&  \textbf{0.8188}& \textbf{0.8526}\\ \Xhline{1.1pt}
    \end{tabular}}

    \label{tab:ablationAAPD}
\end{table}
Label selection directly affects which information from the text will be extracted under the attention mechanism, thus impacting the final label prediction. To validate the effectiveness of using the frequency correlation matrix for label selection, a comparison was made against selecting the same number of labels directly from RoBERTa's predictions without the frequency correlation matrix. Experimental results demonstrate that label selection using the frequency correlation matrix is more effective than directly using RoBERTa's predicted labels. This is understandable, as the relevance of the predicted labels diminishes after a certain quantity. Additionally, RoBERTa's predicted labels do not account for co-occurrence relationships beyond textual semantics. For some labels, the connections are more latent and not directly reflected in the text's semantic content.

The label sequence is sorted based on the frequency of label occurrence across the entire dataset, aiming to introduce the frequency distribution information of labels over the dataset. When expanding the features obtained through the multi-label attention mechanism, this naturally positions high-frequency label features at the forefront, allowing the classifier to inherently pay different levels of attention to labels at different positions.This is because, for the classifier, each feature used for classification, from the head to the tail, consists of features extracted from high-frequency labels to those extracted from low-frequency labels. In comparison to an unordered feature sequence, where the contributions to classification are random, an ordered feature sequence allows for the anticipation that the features towards the front will contribute more. Therefore, as the classifier continuously assesses that the contribution from head labels is greater, the weights assigned to features from head labels will naturally be larger.

This label sequence, enriched with frequency distribution information, is considered a prioritized sequence. For the same label, its meaning varies depending on its position; the further forward it is, the more attention it should receive, and conversely, the further back it is, the less attention it should receive. Therefore, positional information was integrated into the label sequence, enabling different expressions at different positions to achieve dynamic label feature representation.

Experiments across three datasets validated the effectiveness of these approaches. Additionally, an ablation experiment was conducted by removing all the aforementioned designs to verify their overall effect as a reranking module.According to results presented in Table \ref{tab:ablationMAG-CS}, \ref{tab:ablationPubMed}, and \ref{tab:ablationAAPD}, both label sequence ordering and the integration of positional information significantly improved classification outcomes.

\subsubsection{ Effects of the number of labels}
Different choices in the number of labels introduce varying degrees of relevant label information and noise, impacting different datasets and samples differently. The effects of label number selection were investigated through experiments and analysis. Results indicate that in the MAG dataset, 35 labels were optimal (Table \ref{tab:label_num MAG-CS}); in the Mesh dataset, 30 labels were suitable (Table \ref{tab:label_numPubMed}); and in the AAPD dataset, 20 labels were appropriate (Table \ref{tab:label_numAAPD}).

\begin{table}[htb]
\caption{Results with different number of labels on MAG-CS}
\renewcommand{\arraystretch}{1.07}

    \centering
    \resizebox{\linewidth}{!}{
    \begin{tabular}{lccccc} \Xhline{1.1pt} 
         Number of labels&  P@1&  P@3&  P@5&  NDCG@3& NDCG@5\\ \hline
         40&  0.9085&  0.7767&  0.6550&  0.8572& 0.8255\\
         35&  0.9118&  \textbf{0.7811}&  \textbf{0.6591}&  \textbf{0.8620}& \textbf{0.8305}\\ 
         30&  \textbf{0.9125}&  0.7797&  0.6569&  0.8606& 0.8283\\ 
         25&  0.9102&  0.7802&  0.6586&  0.8606& 0.8295\\ 
         20&  0.9055&  0.7722&  0.6053&  0.8527& 0.8207\\ 
 \Xhline{1.1pt}
    \end{tabular}}

    \label{tab:label_num MAG-CS}
\end{table}

\begin{table}[htb]
\caption{Results with different number of labels on PubMed}
\renewcommand{\arraystretch}{1.07}

    \centering
    \resizebox{\linewidth}{!}{
    \begin{tabular}{lccccc} \Xhline{1.1pt} 
         Number of labels&  P@1&  P@3&  P@5&  NDCG@3& NDCG@5
\\ \hline
 35&  0.9275&  0.7805&  0.6567&  \textbf{0.8332}& 0.7735\\
         30&  0.9241&  \textbf{0.7813}&  \textbf{0.6593}&  0.8330& \textbf{0.7749}\\ 
         25&  0.9282&  0.7772&  0.6539&  0.8306& 0.7709\\ 
         20&  0.9275&  0.7765&  0.6511&  0.8300& 0.7688\\ 
         15&  \textbf{0.9293}&  0.7759&  0.6518&  0.8299& 0.7695\\  \Xhline{1.1pt}
    \end{tabular}}

    \label{tab:label_numPubMed}
\end{table}

\begin{table}[htb]
    \caption{Results with different number of labels on AAPD}
\renewcommand{\arraystretch}{1.07}

    \centering
    \resizebox{\linewidth}{!}{
    \begin{tabular}{lccccc} \Xhline{1.1pt} 
         Number of labels&  P@1&  P@3&  P@5&  NDCG@3& NDCG@5
\\ \hline
 30&  0.8480&  0.6130&  \textbf{0.4212}&  0.8102& \textbf{0.8531}\\
         25&  0.8480&  0.6183&  0.4210&  0.8131& 0.8522\\ 
         20&  \textbf{0.8540}&  \textbf{0.6227}&  0.4188&  \textbf{0.8188}& 0.8526\\ 
         15&  0.8480&  0.6177&  0.418&  0.8149& 0.8507\\ 
         10&  0.8350&  0.6117&  0.4172&  0.8041& 0.8434\\  \Xhline{1.1pt}
    \end{tabular}}

        \label{tab:label_numAAPD}
\end{table}

\subsection{Case study}
From an information retrieval perspective, Our model can be viewed as a dual-prediction system from an information retrieval perspective. Initially, text features extracted by RoBERTa are used for preliminary label retrieval within the label space. This initial retrieval expands the relevant label sequence, generating a label information sequence related to the text. Subsequently, an attention mechanism extracts features from the text's word features that are more relevant to these label features for a secondary prediction. We demonstrate the effectiveness of our method by comparing it to the AttentionXML and BERTXML models on the PubMed dataset, illustrating the improvements introduced in the secondary retrieval phase.
\begin{table}[htb]
\caption{Case 1.Extended labels represent additional label information obtained through the correlation matrix. This can be used to analyze why the introduction of additional label information is effective.The prediction results are the top-5 labels predicted by each method.\textcolor{teal}{Green}: Correct predictions. \textcolor{red}{Red}: Incorrect predictions.  \textcolor{orange}{Orange}:Semantics in the text that are directly related to labels.}
\resizebox{\linewidth}{!}{
\begin{tabular}{ll} \Xhline{1.2pt}
Text         & \begin{tabular}[c]{@{}l@{}}gb1 is not a two state folder identification and characterization of an on\\ pathway intermediate the \textcolor{orange}{folding} pathway of the small $\alpha$ $\beta $ protein  gb1\\ has been extensively studied during the past two decades using both\\ theoretical and experimental approaches these studies provided a consensus\\ view that the \textcolor{orange}{protein fold} in a two state manner here we reassessed the\\  \textcolor{orange}{folding} of gb1 both by experiments and simulations and detected the  pre-\\sence of an on pathway intermediate this intermediate has eluded  earlier\\ experimental characterization and is distinct from the collapsed state\\ previously identified using ultrarapid mixing failure to identify the presence\\ of an intermediate affects some of the conclusions that  have been drawn for\\ gb1  a popular model for \textcolor{orange}{protein folding} studies\end{tabular} \\
\hline

Extend labels  & \begin{tabular}[c]{@{}l@{}}`Humans', `Animals', \textcolor{orange}{`Protein Conformation'}, `Protein Structure, Tertiary',\\  `Amino Acid Sequence', `Protein Structure, Secondary', `Protein Binding',\\  \textcolor{orange}{`Kinetics'}, 'Mutation', \textcolor{orange}{`Thermodynamics'}, `Crystallography, X-Ray',\\  `Escherichia coli', `Protein Stability', `Mice', `Protein Multimerization',\\  \textcolor{orange}{`Molecular Dynamics Simulation'}, `Binding Sites', `Protein Denaturation',\\  `Endoplasmic Reticulum', `Amino Acid Substitution', `Hydrophobic and \\Hydrophilic Interactions',  `Protein Transport',  `Hydrogen-Ion Concentration',\\  `Temperature', `Catalytic Domain',  `Sequence Homology, Amino Acid', \\`Saccharomyces cerevisiae', 'HEK293 Cells', `Amyloid'\end{tabular} \\ \hline
Target       & \begin{tabular}[c]{@{}l@{}}`Kinetics', `Protein Folding', `Molecular Dynamics Simulation',\\ `Protein  Conformation',  `Hydrogen-Ion Concentration', `Thermodynamics'\end{tabular}                                                          \\ \hline 
AttentionXML & \begin{tabular}[c]{@{}l@{}}\textcolor{teal}{`Protein Folding'(\ding{52})}, \textcolor{teal}{`Molecular Dynamics Simulation'(\ding{52})}, \\\textcolor{teal}{`Protein }  \textcolor{teal}{Conformation'(\ding{52})},  \textcolor{red}{`Protein Structure,} \textcolor{red}{Secondary'(\ding{56})},\\ \textcolor{red}{`Amino Acid Sequence'(\ding{56})}\end{tabular}                                                                \\ \hline 
BertXML      & \begin{tabular}[c]{@{}l@{}}\textcolor{teal}{`Protein Folding'(\ding{52})}, \textcolor{red}{`Protein Structure, Secondary'(\ding{56})},\\ \textcolor{red}{`Protein Structure}, \textcolor{red}{Tertiary'(\ding{56})}, \textcolor{red}{`Amino Acid Sequence'(\ding{56})},\\  \textcolor{teal}{`Protein Conformation'(\ding{52})}\end{tabular}                                                                              \\ \hline 
LabelCoRank        & \begin{tabular}[c]{@{}l@{}}\textcolor{teal}{`Protein Folding'(\ding{52})}, \textcolor{teal}{`Protein Conformation'(\ding{52})},\\ \textcolor{teal}{`Molecular Dynamics}  \textcolor{teal}{ Simulation'(\ding{52})},  \textcolor{teal}{`Kinetics'(\ding{52})}, \textcolor{teal}{`Thermodynamics'(\ding{52})}\end{tabular}                                                                                            
 \\ \Xhline{1.2pt}\end{tabular}}

  \label{tab:case1}
\end{table}

LabelCoRank is capable of predicting labels associated with features that are difficult to extract from the text. For instance, in Case 1(Table \ref{tab:case1}), the text frequently contains high-frequency keywords such as ``protein'', ``fold'', and ``folding''. Thus, other methods can capture the corresponding features and infer keywords like ``Protein Folding'' and ``Protein Conformation''. However, the correct labels include terms like ``Molecular Dynamics Simulation'', ``Kinetics'', and ``Thermodynamics'', which are scarcely expressed in the text and difficult to capture based solely on the textual content. Our model expands label information beyond the text by using labels predicted in the first pass combined with a frequency relevance matrix. This allows us to obtain labels like ``Kinetics'' and ``Thermodynamics'' in the extended tags. After integrating label embeddings and positional information, the attention mechanism mines text features and incorporates the expanded label features into the final feature expression, enabling the prediction of labels that are difficult to capture from the text alone.

\begin{table}
  \caption{Case 2.Extended labels represent additional label information obtained through the correlation matrix. This can be used to analyze why the introduction of additional label information is effective.The prediction results are the top-5 labels predicted by each method.\textcolor{teal}{Green}: Correct predictions. \textcolor{red}{Red}: Incorrect predictions.  \textcolor{orange}{Orange}:Semantics in the text that are directly related to labels.}
  \resizebox{\linewidth}{!}{
\begin{tabular}{ll} \Xhline{1.1pt} 
Text         & \begin{tabular}[c]{@{}l@{}}the tumor suppressor \textcolor{orange}{apc} differentially regulates multiple $\beta $ \textcolor{orange}{catenin}\\ through the function of axin and cki $\alpha$ during c  \textcolor{orange}{elegans} \textcolor{orange}{asymmetric } \\\textcolor{orange}{ stem cell divisions} abstract the apc tumor suppressor regulates\\ diverse stem \textcolor{orange}{cell} processes including regulation of gene expression\\ through wnt $\beta $ \textcolor{orange}{catenin} signaling  and chromosome stability through\\ microtubule  interactions but how the disparate functions of \textcolor{orange}{apc} are\\ controlled is not well understood acting as part of a wnt signaling\\ pathway $\beta $ \textcolor{orange}{catenin} pathway that controls \textcolor{orange}{asymmetric cell} \textcolor{orange}{division} \\caenorhabditis \textcolor{orange}{elegans} \textcolor{orange}{apc} apr 1 promotes asymmetric nuclear \textcolor{orange}{export}\\ signal of the $\beta $ \textcolor{orange}{catenin} wrm 1 by asymmetrically stabilizing microtubule \\wnt signaling pathway function also dependson a second $\beta $ \textcolor{orange}{catenin} sys\\ 1  which binds to the c \textcolor{orange}{elegans} tcf pop 1 to activate gene expression\\ here we show that apr 1 regulates sys 1 levels in \textcolor{orange}{asymmetric stem cell}\\ \textcolor{orange}{division} in addition to its known role in lowering nuclear levels of wrm 1\\ we demonstrate that sys 1 is also negatively regulated by the c \textcolor{orange}{elegans}\\ homology of \textcolor{orange}{casein}  \textcolor{orange}{kinase} 1 1 $\alpha$ cki$\alpha$ kin 19 we show that kin 19 restricts\\ apr 1 localization  thereby regulating nuclear wrm 1 finally the polarity of\\ apr 1 cortical  localization is controlled by pry 1 c \textcolor{orange}{elegans} axin such that\\ pry 1 controls the polarity of both sys 1 and wrm 1 asymmetries we\\ propose a model whereby wnt signaling pathway through cki$\alpha$ regulates\\ the function of two distinct  pools of \textcolor{orange}{apc} one \textcolor{orange}{apc} pool negatively regulates\\ sys 1 whereas the second pool stabilizes microtubule and promotes wrm 1\\ nuclear export signal\end{tabular} \\

\hline  
Extend labels  & \begin{tabular}[c]{@{}l@{}}\textcolor{orange}{`Mice'}, `Humans', \textcolor{orange}{`Mice, Inbred C57BL'}, \textcolor{orange}{`Rats'}, `Signal Transduction', \\`Mice, Knockout', `Disease Models, Animal', \textcolor{orange}{`Cells, Cultured'},\\ \textcolor{orange}{`Cell Line, Tumor'},  \textcolor{orange}{`Cell Line'}, `Neurons', `Mice, Transgenic', \\`Rats, Sprague-Dawley', \textcolor{orange}{`Cell } 
 \textcolor{orange}{Proliferation'},  `Gene Expression Regulation', \\`Mutation', \textcolor{orange}{`Cell Differentiation'}, \textcolor{orange}{`Protein Binding'},  `Mice, Inbred \\ BALB C', `Phosphorylation', `Time Factors', `Brain', `Amino Acid Sequence', \\`HEK293 Cells', \textcolor{orange}{`Cell Movement'}, `Rats, Wistar',  `Immunohistochemistry',\\`Macrophages', `Liver'\end{tabular}\\ \hline
Target       & \begin{tabular}[c]{@{}l@{}}`Protein Kinases', `Adenomatous Polyposis Coli Protein', `Animals',\\ `Protein Transport',  `Microtubules', `Stem Cells', `Asymmetric Cell Division',\\  `Cell Nucleus', `Active Transport, Cell Nucleus', `Wnt Signaling Pathway',  \\`Cell Polarity'\end{tabular}                                                                                \\ \hline
AttentionXML & \begin{tabular}[c]{@{}l@{}} \textcolor{teal}{`Animals'(\ding{52})}, \textcolor{red}{`Axin Protein(\ding{56})'}, \textcolor{teal}{`Microtubules'(\ding{52})}, \textcolor{red}{`Cell Division'(\ding{56})},\\ \textcolor{teal}{`Cell Nucleus'(\ding{52})}    \end{tabular}                                                                                      \\ \hline 
BertXML      & \begin{tabular}[c]{@{}l@{}}\textcolor{teal}{`Animals'(\ding{52})}, \textcolor{red}{`Protein-Serine-Threonine Kinases'(\ding{56})}, \textcolor{red}{`Mutation'(\ding{56})},\\  \textcolor{red}{`RNA Interference'(\ding{56})}, \textcolor{red} {`Gene Expression }\textcolor{red}{Regulation, Developmental'(\ding{56})}\end{tabular}                                                                                                                                        \\ \hline 
LabelCoRank        & \begin{tabular}[c]{@{}l@{}}\textcolor{teal}{`Animals'(\ding{52})}, \textcolor{teal}{`Asymmetric Cell Division'(\ding{52})},\\ \textcolor{teal}{`Adenomatous Polyposis} \textcolor{teal}{Coli Protein'(\ding{52})},  \textcolor{teal}{`Protein Transport'(\ding{52})},\\ \textcolor{teal}{`Stem Cells'(\ding{52})}\end{tabular}                                                                                                       
 \\ \Xhline{1.1pt}\end{tabular}}

    \label{tab:case2}
\end{table}

LabelCoRank enhances text feature capture specific to labels. During the process of label sequence expansion, it is not necessary to directly obtain the correct labels; acquiring semantically similar labels also aids in label prediction. For example, in Case 2(Table \ref{tab:case2}),  some keywords appearing in the text do not directly associate with labels but are linked through similar semantics, such as ``catenin'' and ``protein'' or ``elegans'' and ``animals''. Through label expansion, relevant labels may not be directly acquired, but more semantically related labels are obtained.  For example, in predicting labels such as ``Cell Nucleus'' and ``Cell Polarity'', similar features of labels such as ``Cells, Cultured'', ``Cell Line, Tumor'', ``Cell Line'', and ``Cell Differentiation'' are utilized through a label attention mechanism. This allows more features related to ``Cell'' to be extracted, thus aiding label prediction.

\section{CONCLUSION}
This paper introduces LabelCoRank, a model that integrates label information by Label Reranking method. The principal mechanism of this model is to expand the labels related to the text by using label frequency correlation information, which allows for the extraction of more effective features targeted at labels through the operation of a label attention mechanism. Additionally, the integration of label frequency distribution information into the label feature sequence via sorting and positional information helps the model to capture more significant features for classification. We compared our model with current advanced models on three public datasets: MAG-CS, PubMed, and AAPD, demonstrating its effectiveness. The efficacy of the modules was validated through ablation studies, and case analyses illustrated how the introduction of label information via our method positively influences the prediction results.

\section{Competing interests}
No competing interest is declared.



\section{Acknowledgments}
This work was supported by National High-End Foreign Experts Program(H20240916) and National Natural Science Foundation of China(52374165,52304185).

\bibliographystyle{plain}
\bibliography{reference}

\begin{thebibliography}{10}

\bibitem{bai2018empirical}
Shaojie Bai, J.~Zico Kolter, and Vladlen Koltun.
\newblock An empirical evaluation of generic convolutional and recurrent networks for sequence modeling.
\newblock {\em ArXiv}, abs/1803.01271, 2018.

\bibitem{pmlr-v97-byrd19a}
Jonathon Byrd and Zachary Lipton.
\newblock What is the effect of importance weighting in deep learning?
\newblock In Kamalika Chaudhuri and Ruslan Salakhutdinov, editors, {\em Proceedings of the 36th International Conference on Machine Learning}, volume~97 of {\em Proceedings of Machine Learning Research}, pages 872--881. PMLR, 09--15 Jun 2019.

\bibitem{Chang_Wang_Peng_2024}
Ying-Ying Chang, Wei-Yao Wang, and Wen-Chih Peng.
\newblock Sega: Preference-aware self-contrastive learning with prompts for anomalous user detection on twitter.
\newblock {\em Proceedings of the AAAI Conference on Artificial Intelligence}, 38(1):30--37, Mar. 2024.

\bibitem{chen2022survey}
Xiaolong Chen, Jieren Cheng, Jingxin Liu, Wenghang Xu, Shuai Hua, Zhu Tang, and Victor~S Sheng.
\newblock A survey of multi-label text classification based on deep learning.
\newblock In {\em International Conference on Adaptive and Intelligent Systems}, pages 443--456. Springer, 2022.

\bibitem{chen2015convolutional}
Yahui Chen.
\newblock Convolutional neural network for sentence classification.
\newblock Master's thesis, University of Waterloo, 2015.

\bibitem{Cui_Jia_2024}
Chaoqun Cui and Caiyan Jia.
\newblock Propagation tree is not deep: Adaptive graph contrastive learning approach for rumor detection.
\newblock {\em Proceedings of the AAAI Conference on Artificial Intelligence}, 38(1):73--81, Mar. 2024.

\bibitem{8953804}
Yin Cui, Menglin Jia, Tsung-Yi Lin, Yang Song, and Serge Belongie.
\newblock Class-balanced loss based on effective number of samples.
\newblock In {\em 2019 IEEE/CVF Conference on Computer Vision and Pattern Recognition (CVPR)}, pages 9260--9269, 2019.

\bibitem{devlin-etal-2019-bert}
Jacob Devlin, Ming-Wei Chang, Kenton Lee, and Kristina Toutanova.
\newblock {BERT}: Pre-training of deep bidirectional transformers for language understanding.
\newblock In Jill Burstein, Christy Doran, and Thamar Solorio, editors, {\em Proceedings of the 2019 Conference of the North {A}merican Chapter of the Association for Computational Linguistics: Human Language Technologies, Volume 1 (Long and Short Papers)}, pages 4171--4186, Minneapolis, Minnesota, June 2019. Association for Computational Linguistics.

\bibitem{10.1145/3097983.3098016}
Yuxiao Dong, Hao Ma, Zhihong Shen, and Kuansan Wang.
\newblock A century of science: Globalization of scientific collaborations, citations, and innovations.
\newblock In {\em Proceedings of the 23rd ACM SIGKDD International Conference on Knowledge Discovery and Data Mining}, KDD '17, page 1437–1446, New York, NY, USA, 2017. Association for Computing Machinery.

\bibitem{Goudjil2018}
Mohamed Goudjil, Mouloud Koudil, Mouldi Bedda, and Noureddine Ghoggali.
\newblock A novel active learning method using svm for text classification.
\newblock {\em International Journal of Automation and Computing}, 15(3):290--298, Jun 2018.

\bibitem{GU2023110025}
Tiquan Gu, Hui Zhao, Zhenzhen He, Min Li, and Di~Ying.
\newblock Integrating external knowledge into aspect-based sentiment analysis using graph neural network.
\newblock {\em Knowledge-Based Systems}, 259:110025, 2023.

\bibitem{Guo_Yang_Liu_Bai_Wang_Li_Zheng_Zhang_Peng_Tian_2024}
Hongcheng Guo, Jian Yang, Jiaheng Liu, Jiaqi Bai, Boyang Wang, Zhoujun Li, Tieqiao Zheng, Bo~Zhang, Junran Peng, and Qi~Tian.
\newblock Logformer: A pre-train and tuning pipeline for log anomaly detection.
\newblock {\em Proceedings of the AAAI Conference on Artificial Intelligence}, 38(1):135--143, Mar. 2024.

\bibitem{guo-etal-2019-star}
Qipeng Guo, Xipeng Qiu, Pengfei Liu, Yunfan Shao, Xiangyang Xue, and Zheng Zhang.
\newblock Star-transformer.
\newblock In Jill Burstein, Christy Doran, and Thamar Solorio, editors, {\em Proceedings of the 2019 Conference of the North {A}merican Chapter of the Association for Computational Linguistics: Human Language Technologies, Volume 1 (Long and Short Papers)}, pages 1315--1325, Minneapolis, Minnesota, June 2019. Association for Computational Linguistics.

\bibitem{10.1145/3511808.3557632}
Sangwoo Han, Eunseong Choi, Chan Lim, Hyunjung Shim, and Jongwuk Lee.
\newblock Long-tail mixup for extreme multi-label classification.
\newblock In {\em Proceedings of the 31st ACM International Conference on Information \& Knowledge Management}, CIKM '22, page 3998–4002, New York, NY, USA, 2022. Association for Computing Machinery.

\bibitem{huang-etal-2021-balancing}
Yi~Huang, Buse Giledereli, Abdullatif K{\"o}ksal, Arzucan {\"O}zg{\"u}r, and Elif Ozkirimli.
\newblock Balancing methods for multi-label text classification with long-tailed class distribution.
\newblock In Marie-Francine Moens, Xuanjing Huang, Lucia Specia, and Scott Wen-tau Yih, editors, {\em Proceedings of the 2021 Conference on Empirical Methods in Natural Language Processing}, pages 8153--8161, Online and Punta Cana, Dominican Republic, November 2021. Association for Computational Linguistics.

\bibitem{JAYALAKSHMI2023103351}
N~Jayalakshmi, V~Sangeeta, and Appala~Srinuvasu Muttipati.
\newblock Taylor horse herd optimized deep fuzzy clustering and laplace based k-nearest neighbor for web page recommendation.
\newblock {\em Advances in Engineering Software}, 175:103351, 2023.

\bibitem{Jiang2023}
Weili Jiang, Kangneng Zhou, Chenchen Xiong, Guodong Du, Chubin Ou, and Junpeng Zhang.
\newblock Kscb: a novel unsupervised method for text sentiment analysis.
\newblock {\em Applied Intelligence}, 53(1):301--311, Jan 2023.

\bibitem{kang2019decoupling}
Bingyi Kang, Saining Xie, Marcus Rohrbach, Zhicheng Yan, Albert Gordo, Jiashi Feng, and Yannis Kalantidis.
\newblock Decoupling representation and classifier for long-tailed recognition.
\newblock In {\em Eighth International Conference on Learning Representations (ICLR)}, 2020.

\bibitem{9430169}
Anwesha Law and Ashish Ghosh.
\newblock Multi-label classification using binary tree of classifiers.
\newblock {\em IEEE Transactions on Emerging Topics in Computational Intelligence}, 6(3):677--689, 2022.

\bibitem{li-etal-2023-towards-better}
Fangfang Li, Puzhen Su, Junwen Duan, and Weidong Xiao.
\newblock Towards better representations for multi-label text classification with multi-granularity information.
\newblock In Houda Bouamor, Juan Pino, and Kalika Bali, editors, {\em Findings of the Association for Computational Linguistics: EMNLP 2023}, pages 9470--9480, Singapore, December 2023. Association for Computational Linguistics.

\bibitem{li-etal-2022-ligcn}
Irene Li, Aosong Feng, Hao Wu, Tianxiao Li, Toyotaro Suzumura, and Ruihai Dong.
\newblock {L}i{GCN}: Label-interpretable graph convolutional networks for multi-label text classification.
\newblock In Lingfei Wu, Bang Liu, Rada Mihalcea, Jian Pei, Yue Zhang, and Yunyao Li, editors, {\em Proceedings of the 2nd Workshop on Deep Learning on Graphs for Natural Language Processing (DLG4NLP 2022)}, pages 60--70, Seattle, Washington, July 2022. Association for Computational Linguistics.

\bibitem{10.1145/3077136.3080834}
Jingzhou Liu, Wei-Cheng Chang, Yuexin Wu, and Yiming Yang.
\newblock Deep learning for extreme multi-label text classification.
\newblock In {\em Proceedings of the 40th International ACM SIGIR Conference on Research and Development in Information Retrieval}, SIGIR '17, page 115–124, New York, NY, USA, 2017. Association for Computing Machinery.

\bibitem{10.5555/3060832.3061023}
Pengfei Liu, Xipeng Qiu, and Xuanjing Huang.
\newblock Recurrent neural network for text classification with multi-task learning.
\newblock In {\em Proceedings of the Twenty-Fifth International Joint Conference on Artificial Intelligence}, IJCAI'16, page 2873–2879. AAAI Press, 2016.

\bibitem{4717268}
Xu-Ying Liu, Jianxin Wu, and Zhi-Hua Zhou.
\newblock Exploratory undersampling for class-imbalance learning.
\newblock {\em IEEE Transactions on Systems, Man, and Cybernetics, Part B (Cybernetics)}, 39(2):539--550, 2009.

\bibitem{liu2019roberta}
Yinhan Liu, Myle Ott, Naman Goyal, Jingfei Du, Mandar Joshi, Danqi Chen, Omer Levy, Mike Lewis, Luke Zettlemoyer, and Veselin Stoyanov.
\newblock Roberta: A robustly optimized bert pretraining approach.
\newblock {\em ArXiv}, abs/1907.11692, 2019.

\bibitem{Lu2011-gz}
Zhiyong Lu.
\newblock {PubMed} and beyond: a survey of web tools for searching biomedical literature.
\newblock {\em Database (Oxford)}, 2011(0):baq036, January 2011.

\bibitem{ma-etal-2021-label}
Qianwen Ma, Chunyuan Yuan, Wei Zhou, and Songlin Hu.
\newblock Label-specific dual graph neural network for multi-label text classification.
\newblock In Chengqing Zong, Fei Xia, Wenjie Li, and Roberto Navigli, editors, {\em Proceedings of the 59th Annual Meeting of the Association for Computational Linguistics and the 11th International Joint Conference on Natural Language Processing (Volume 1: Long Papers)}, pages 3855--3864, Online, August 2021. Association for Computational Linguistics.

\bibitem{10.1007/978-3-030-01216-8_12}
Dhruv Mahajan, Ross Girshick, Vignesh Ramanathan, Kaiming He, Manohar Paluri, Yixuan Li, Ashwin Bharambe, and Laurens van~der Maaten.
\newblock Exploring the limits of weakly supervised pretraining.
\newblock In Vittorio Ferrari, Martial Hebert, Cristian Sminchisescu, and Yair Weiss, editors, {\em Computer Vision -- ECCV 2018}, pages 185--201, Cham, 2018. Springer International Publishing.

\bibitem{mao2017mesh}
Yuqing Mao and Zhiyong Lu.
\newblock Mesh now: automatic mesh indexing at pubmed scale via learning to rank.
\newblock {\em Journal of biomedical semantics}, 8:1--9, 2017.

\bibitem{Nassiri2023}
Khalid Nassiri and Moulay Akhloufi.
\newblock Transformer models used for text-based question answering systems.
\newblock {\em Applied Intelligence}, 53(9):10602--10635, May 2023.

\bibitem{pal2020multi}
Ankit Pal, Muru Selvakumar, and Malaikannan Sankarasubbu.
\newblock Multi-label text classification using attention-based graph neural network.
\newblock {\em ArXiv}, abs/2003.11644, 2020.

\bibitem{8117734}
Anjuman Prabhat and Vikas Khullar.
\newblock Sentiment classification on big data using naïve bayes and logistic regression.
\newblock In {\em 2017 International Conference on Computer Communication and Informatics (ICCCI)}, pages 1--5, 2017.

\bibitem{2018Parabel}
Yashoteja Prabhu, Anil Kag, Shrutendra Harsola, Rahul Agrawal, and Manik Varma.
\newblock Parabel: Partitioned label trees for extreme classification with application to dynamic search advertising.
\newblock In {\em the 2018 World Wide Web Conference}, 2018.

\bibitem{ROSEBERRY202110}
Martha Roseberry, Bartosz Krawczyk, Youcef Djenouri, and Alberto Cano.
\newblock Self-adjusting k nearest neighbors for continual learning from multi-label drifting data streams.
\newblock {\em Neurocomputing}, 442:10--25, 2021.

\bibitem{SUN2024111878}
Guoying Sun, Yanan Cheng, Fangzhou Dong, Luhua Wang, Dong Zhao, Zhaoxin Zhang, and Xiaojun Tong.
\newblock Multi-label text classification model integrating label attention and historical attention.
\newblock {\em Knowledge-Based Systems}, 296:111878, 2024.

\bibitem{NIPS2017_3f5ee243}
Ashish Vaswani, Noam Shazeer, Niki Parmar, Jakob Uszkoreit, Llion Jones, Aidan~N Gomez, \L~ukasz Kaiser, and Illia Polosukhin.
\newblock Attention is all you need.
\newblock In I.~Guyon, U.~Von Luxburg, S.~Bengio, H.~Wallach, R.~Fergus, S.~Vishwanathan, and R.~Garnett, editors, {\em Advances in Neural Information Processing Systems}, volume~30. Curran Associates, Inc., 2017.

\bibitem{Vu2023}
Huy-The Vu, Minh-Tien Nguyen, Van-Chien Nguyen, Minh-Hieu Pham, Van-Quyet Nguyen, and Van-Hau Nguyen.
\newblock Label-representative graph convolutional network for multi-label text classification.
\newblock {\em Applied Intelligence}, 53(12):14759--14774, Jun 2023.

\bibitem{10.1145/3643853}
Haobo Wang, Cheng Peng, Hede Dong, Lei Feng, Weiwei Liu, Tianlei Hu, Ke~Chen, and Gang Chen.
\newblock On the value of head labels in multi-label text classification.
\newblock {\em ACM Trans. Knowl. Discov. Data}, 18(5), March 2024.

\bibitem{10.1162/qss_a_00021}
Kuansan Wang, Zhihong Shen, Chiyuan Huang, Chieh-Han Wu, Yuxiao Dong, and Anshul Kanakia.
\newblock {Microsoft Academic Graph: When experts are not enough}.
\newblock {\em Quantitative Science Studies}, 1(1):396--413, 02 2020.

\bibitem{WANG2023161}
Qing Wang, Jia Zhu, Hongji Shu, Kwame~Omono Asamoah, Jianyang Shi, and Cong Zhou.
\newblock Gudn: A novel guide network with label reinforcement strategy for extreme multi-label text classification.
\newblock {\em Journal of King Saud University - Computer and Information Sciences}, 35(4):161--171, 2023.

\bibitem{wang-etal-2022-kenmesh}
Xindi Wang, Robert Mercer, and Frank Rudzicz.
\newblock {K}en{M}e{SH}: Knowledge-enhanced end-to-end biomedical text labelling.
\newblock In Smaranda Muresan, Preslav Nakov, and Aline Villavicencio, editors, {\em Proceedings of the 60th Annual Meeting of the Association for Computational Linguistics (Volume 1: Long Papers)}, pages 2941--2951, Dublin, Ireland, May 2022. Association for Computational Linguistics.

\bibitem{Wang_Chu_Ouyang_Wang_Hao_Shen_Gu_Xue_Zhang_Cui_Li_Zhou_Li_2024}
Yan Wang, Zhixuan Chu, Xin Ouyang, Simeng Wang, Hongyan Hao, Yue Shen, Jinjie Gu, Siqiao Xue, James Zhang, Qing Cui, Longfei Li, Jun Zhou, and Sheng Li.
\newblock Llmrg: Improving recommendations through large language model reasoning graphs.
\newblock {\em Proceedings of the AAAI Conference on Artificial Intelligence}, 38(17):19189--19196, Mar. 2024.

\bibitem{Wang_2019_ICCV}
Yiru Wang, Weihao Gan, Jie Yang, Wei Wu, and Junjie Yan.
\newblock Dynamic curriculum learning for imbalanced data classification.
\newblock In {\em Proceedings of the IEEE/CVF International Conference on Computer Vision (ICCV)}, October 2019.

\bibitem{Wu_Qiu_Zheng_Zhu_Chen_2024}
Likang Wu, Zhaopeng Qiu, Zhi Zheng, Hengshu Zhu, and Enhong Chen.
\newblock Exploring large language model for graph data understanding in online job recommendations.
\newblock {\em Proceedings of the AAAI Conference on Artificial Intelligence}, 38(8):9178--9186, Mar. 2024.

\bibitem{Loss1}
Tong Wu, Qingqiu Huang, Ziwei Liu, Yu~Wang, and Dahua Lin.
\newblock Distribution-balanced loss for multi-label classification in long-tailed datasets.
\newblock In {\em Computer Vision – ECCV 2020: 16th European Conference, Glasgow, UK, August 23–28, 2020, Proceedings, Part IV}, page 162–178, Berlin, Heidelberg, 2020. Springer-Verlag.

\bibitem{xiao-etal-2019-label}
Lin Xiao, Xin Huang, Boli Chen, and Liping Jing.
\newblock Label-specific document representation for multi-label text classification.
\newblock In Kentaro Inui, Jing Jiang, Vincent Ng, and Xiaojun Wan, editors, {\em Proceedings of the 2019 Conference on Empirical Methods in Natural Language Processing and the 9th International Joint Conference on Natural Language Processing (EMNLP-IJCNLP)}, pages 466--475, Hong Kong, China, November 2019. Association for Computational Linguistics.

\bibitem{xiao2021does}
Lin Xiao, Xiangliang Zhang, Liping Jing, Chi Huang, and Mingyang Song.
\newblock Does head label help for long-tailed multi-label text classification.
\newblock In {\em Proceedings of the AAAI conference on artificial intelligence}, volume~35, pages 14103--14111, 2021.

\bibitem{xu-etal-2021-hierarchical}
Linli Xu, Sijie Teng, Ruoyu Zhao, Junliang Guo, Chi Xiao, Deqiang Jiang, and Bo~Ren.
\newblock Hierarchical multi-label text classification with horizontal and vertical category correlations.
\newblock In Marie-Francine Moens, Xuanjing Huang, Lucia Specia, and Scott Wen-tau Yih, editors, {\em Proceedings of the 2021 Conference on Empirical Methods in Natural Language Processing}, pages 2459--2468, Online and Punta Cana, Dominican Republic, November 2021. Association for Computational Linguistics.

\bibitem{10.1145/3394486.3403151}
Guangxu Xun, Kishlay Jha, Jianhui Sun, and Aidong Zhang.
\newblock Correlation networks for extreme multi-label text classification.
\newblock In {\em Proceedings of the 26th ACM SIGKDD International Conference on Knowledge Discovery \& Data Mining}, KDD '20, page 1074–1082, New York, NY, USA, 2020. Association for Computing Machinery.

\bibitem{10.1093/bioinformatics/btz142}
Guangxu Xun, Kishlay Jha, Ye~Yuan, Yaqing Wang, and Aidong Zhang.
\newblock {MeSHProbeNet: a self-attentive probe net for MeSH indexing}.
\newblock {\em Bioinformatics}, 35(19):3794--3802, 03 2019.

\bibitem{yang-etal-2018-sgm}
Pengcheng Yang, Xu~Sun, Wei Li, Shuming Ma, Wei Wu, and Houfeng Wang.
\newblock {SGM}: Sequence generation model for multi-label classification.
\newblock In Emily~M. Bender, Leon Derczynski, and Pierre Isabelle, editors, {\em Proceedings of the 27th International Conference on Computational Linguistics}, pages 3915--3926, Santa Fe, New Mexico, USA, August 2018. Association for Computational Linguistics.

\bibitem{NEURIPS2019_dc6a7e65}
Zhilin Yang, Zihang Dai, Yiming Yang, Jaime Carbonell, Russ~R Salakhutdinov, and Quoc~V Le.
\newblock Xlnet: Generalized autoregressive pretraining for language understanding.
\newblock In H.~Wallach, H.~Larochelle, A.~Beygelzimer, F.~d\textquotesingle Alch\'{e}-Buc, E.~Fox, and R.~Garnett, editors, {\em Advances in Neural Information Processing Systems}, volume~32. Curran Associates, Inc., 2019.

\bibitem{yang-etal-2016-hierarchical}
Zichao Yang, Diyi Yang, Chris Dyer, Xiaodong He, Alex Smola, and Eduard Hovy.
\newblock Hierarchical attention networks for document classification.
\newblock In Kevin Knight, Ani Nenkova, and Owen Rambow, editors, {\em Proceedings of the 2016 Conference of the North {A}merican Chapter of the Association for Computational Linguistics: Human Language Technologies}, pages 1480--1489, San Diego, California, June 2016. Association for Computational Linguistics.

\bibitem{yinhan2019roberta}
Liu Yinhan, Ott Myle, Goyal Naman, Du~Jingfei, Joshi Mandar, Chen Danqi, Levy Omer, and Lewis Mike.
\newblock Roberta: A robustly optimized bert pretraining approach (2019).
\newblock {\em arXiv preprint arXiv:1907.11692}, pages 1--13, 2019.

\bibitem{NEURIPS2019_9e6a921f}
Ronghui You, Zihan Zhang, Ziye Wang, Suyang Dai, Hiroshi Mamitsuka, and Shanfeng Zhu.
\newblock Attentionxml: Label tree-based attention-aware deep model for high-performance extreme multi-label text classification.
\newblock In H.~Wallach, H.~Larochelle, A.~Beygelzimer, F.~d\textquotesingle Alch\'{e}-Buc, E.~Fox, and R.~Garnett, editors, {\em Advances in Neural Information Processing Systems}, volume~32. Curran Associates, Inc., 2019.

\bibitem{ZENG2024111303}
Delong Zeng, Enze Zha, Jiayi Kuang, and Ying Shen.
\newblock Multi-label text classification based on semantic-sensitive graph convolutional network.
\newblock {\em Knowledge-Based Systems}, 284:111303, 2024.

\bibitem{zhang-etal-2021-enhancing}
Ximing Zhang, Qian-Wen Zhang, Zhao Yan, Ruifang Liu, and Yunbo Cao.
\newblock Enhancing label correlation feedback in multi-label text classification via multi-task learning.
\newblock In Chengqing Zong, Fei Xia, Wenjie Li, and Roberto Navigli, editors, {\em Findings of the Association for Computational Linguistics: ACL-IJCNLP 2021}, pages 1190--1200, Online, August 2021. Association for Computational Linguistics.

\bibitem{10.1145/3442381.3449979}
Yu~Zhang, Zhihong Shen, Yuxiao Dong, Kuansan Wang, and Jiawei Han.
\newblock Match: Metadata-aware text classification in a large hierarchy.
\newblock In {\em Proceedings of the Web Conference 2021}, WWW '21, page 3246–3257, New York, NY, USA, 2021. Association for Computing Machinery.

\bibitem{Zhang2021LearningFS}
Zizhao Zhang and Tomas Pfister.
\newblock Learning fast sample re-weighting without reward data.
\newblock {\em 2021 IEEE/CVF International Conference on Computer Vision (ICCV)}, pages 705--714, 2021.

\bibitem{ZHAO2024109842}
Wei Zhao and Hong Zhao.
\newblock Hierarchical long-tailed classification based on multi-granularity knowledge transfer driven by multi-scale feature fusion.
\newblock {\em Pattern Recognition}, 145:109842, 2024.

\end{thebibliography}



\end{document}